\documentclass{article}
\PassOptionsToPackage{numbers, compress}{natbib}
\usepackage[preprint]{neurips_2021}
\usepackage[utf8]{inputenc} 
\usepackage[T1]{fontenc}    
\usepackage{hyperref}       
\usepackage{url}            
\usepackage{booktabs}       
\usepackage{amsfonts}       
\usepackage{nicefrac}       
\usepackage{microtype}      
\usepackage{xcolor}         
\usepackage[pdftex]{graphicx}      
\usepackage{caption}
\usepackage{subcaption}
\usepackage{siunitx}
\newtheorem{assumption}{Assumption}

\definecolor{CommentOrange}{rgb}{0.8,0.4,0.0}
\newcommand{\pop}{\mathcal{P}}

\title{Faster Improvement Rate Population Based Training}

\author{%
  Valentin Dalibard \qquad Max Jaderberg\\
  DeepMind, London, \\
  \texttt{\{vdalibard, jaderberg\}@deepmind.com}
  
}

\begin{document}

\maketitle

\begin{abstract}
  The successful training of neural networks typically involves careful and time consuming hyperparameter tuning. Population Based Training (PBT) has recently been proposed to automate this process. PBT trains a population of neural networks concurrently, frequently mutating their hyperparameters throughout their training. However, the decision mechanisms of PBT  are greedy and favour short-term improvements which can, in some cases, lead to poor long-term performance. This paper presents Faster Improvement Rate PBT (FIRE PBT) which addresses this problem. Our method is guided by an assumption: given two neural networks with similar performance and training with similar hyperparameters, the network showing the faster rate of improvement will lead to a better final performance. Using this, we derive a novel fitness metric and use it to make some of the population members focus on long-term performance. Our experiments show that FIRE PBT is able to outperform PBT on the ImageNet benchmark and match the performance of networks that were trained with a hand-tuned learning rate schedule. We apply FIRE PBT to reinforcement learning tasks and show that it leads to faster learning and higher final performance than both PBT and random hyperparameter search.
\end{abstract}

\section{Introduction}
In order to successfully train neural networks, it is crucial that their hyperparameters are carefully tuned. This is a time consuming task and in recent years, the field of AutoML \cite{automl} has emerged to design methods automating this process \cite{snoek2012practical, hutter2011sequential, bergstra2011algorithms, li2016hyperband}. Population based training (PBT) is one such method. In PBT, a population of workers concurrently train their respective neural networks. The workers regularly explore the hyperparameter space by mutating their hyperparameters while training. With the same amount of resources, PBT can outperform random hyperparameter search on many important problems \cite{PBT, impala, ctf}.

\begin{figure}
  \raggedleft
  \includegraphics[width=0.92\linewidth]{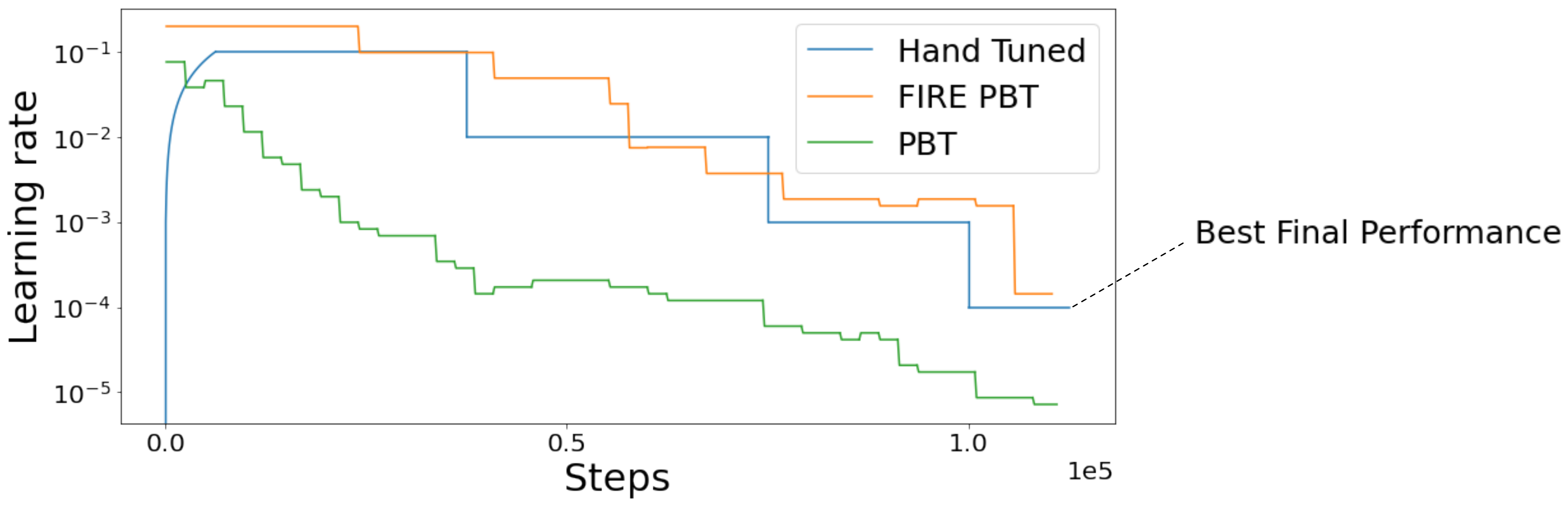}
  \caption{Hand-tuned learning rate schedule for ImageNet training \cite{goyal} compared with the best schedules found by PBT and FIRE PBT after similar number of steps. As shown in Section \ref{sec:experiments}, the average test accuracy scored by executing a schedule found by PBT is 71.7, while FIRE PBT and the hand-tuned schedule score 76.5 and 76.6 respectively.}
  \label{fig:fig1}
\end{figure}

However, PBT's decisions favour immediate improvements in performance and can sometimes lead to unsatisfactory  solutions later in the training process. For example, this is often the case with the learning rate hyperparameter. The best learning rate schedules will often keep the learning rate high for a large fraction of training before reducing it. However, decaying the learning rate during training often results in an immediate boost in performance. When PBT is made to optimise the learning rate, it will often greedily reduce it early on in training (see Figure \ref{fig:fig1}) resulting in a poor final performance. 

This paper presents Fastest Improvement Rate (FIRE) PBT which addresses this problem. In FIRE PBT, disjoints \emph{sub-populations} $\pop_1, \pop_2 \dots \pop_n$ each train with their own dedicated fitness function. The fitness functions are designed so that each sub-population converges on hyperparameters corresponding to a different section of a good hyperparameter schedule. They are based on a novel criterion which measures the improvement rate of training curves. 

Our experiments show that FIRE PBT outperforms PBT and matches the performance of networks that were trained with a hand-tuned learning rate schedule on the ImageNet benchmark \cite{imagenet}. We also apply FIRE PBT to a V-MPO agent \cite{vmpo} on a selection of OpenAI Gym tasks \cite{openaigym} and show that it leads to faster learning and higher performance than PBT or random hyperparameter search.

We review the mechanisms of PBT in Section \ref{sec:pbt}. Section \ref{sec:firepbt} introduces FIRE PBT. We compare the performance of Random Search (RS), PBT and FIRE PBT in Section \ref{sec:experiments}. Finally, we review the related work in Section \ref{sec:relatedwork} and conclude in Section \ref{sec:conclusion}.

\section{Population based training}
\label{sec:pbt}

This section reviews the core mechanisms of Population based training as described by Jaderberg et al. \cite{PBT}. PBT is an optimisation method for the training of neural networks. The goal is to maximize an objective function $\mathcal{Q}$, such as classification accuracy on a validation set. 

In PBT, a population of \emph{workers} (also called \emph{population members}) concurrently train their respective neural networks, each with their own hyperparameters. At regular intervals, each population member compares its evaluation of $\mathcal{Q}$, called \emph{fitness}, with the rest of the population. If a population member $\rho$ has a low fitness compared with its peers, it undergoes an \emph{exploit-and-explore} step. In the exploit step, $\rho$ discards its own state and copies the neural network weights and hyperparameters of a better performing population member. In the explore step, $\rho$ mutates the copied hyperparameters.

Compared with sequential hyperparameter optimisation methods such as Bayesian optimisation~\cite{bayesopt}, PBT uses parallel training to complete within the wall clock time of a single learning process. The hyperparameters will be optimised concurrently to the training of the neural network. Compared with parallel optimisation such as random hyperparameter search, PBT tends to yield better performance~\cite{PBT}. 

However, PBT uses a greedy process which can lead to low performance later in training. For example, this can occur when PBT is made to optimise the learning rate using the current validation accuracy as fitness. Because decaying the learning rate results in an immediate improvement in performance, PBT will greedily decay the learning rate from the start of the experiment, as shown in Figure \ref{fig:fig1}. However, good hand-tuned learning rate schedules will often keep the learning rate high for a significant fraction of training before decaying it. As a result, the performance of networks trained with PBT will be lower than the one obtained by a good hand-tuned schedule.

\section{FIRE PBT}
\label{sec:firepbt}

\begin{figure}
  \centering
  \includegraphics[width=\linewidth]{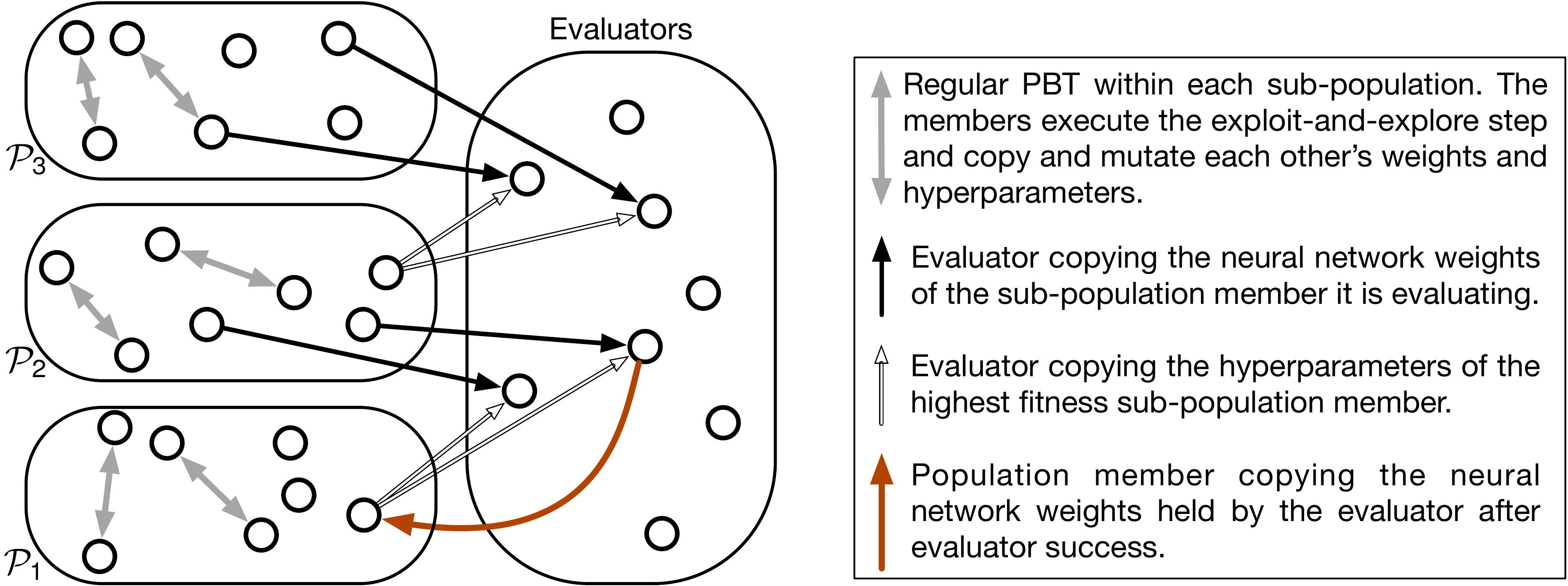}
  \caption{The mechanisms of FIRE PBT. Each circle represents a population member. The arrows indicate the copying of neural network weights and/or hyperparameters between members.}
  \label{fig:mechanisms}
\end{figure}

We now present the mechanisms of FIRE PBT, which are summarised in Figure \ref{fig:mechanisms}. Like in PBT, a FIRE PBT experiment consists of many workers training neural networks concurrently. Each worker $\rho$ holds a pair $(\theta_\rho, h_\rho)$ where $\theta_\rho$ are neural network weights and $h_\rho$ are training hyperparameters. 

We divide the workers into two groups. $\pop$ = $\{\rho_i\}_{i=1}^P$, which we call the population members, and $\mathcal{H}$ = $\{\eta_i\}_{i=1}^E$, which we refer to as the evaluators. The population $\pop$ is further divided into multiple disjoint sub-populations $\pop_1, \pop_2 \dots \pop_n$.

Within each sub-population $\pop_i$, the population members internally run regular PBT. Members have an associated fitness and the ones with low fitness perform the exploit-and-explore step with other members of $\pop_i$. What distinguishes sub-populations from one another is the fitness function they use.

Sub-population $\pop_1$ is greedy and uses $\mathcal{Q}$ as its fitness, similar to regular PBT. For example, $\mathcal{Q}$ could be the classification accuracy on a validation set. All other sub-populations $\pop_{i}$ with $i>1$ are \emph{parent} sub-populations and behave differently. Their fitness encourages them to produce neural network weights which reach high values of $\mathcal{Q}$ when trained with the hyperparameters used in their child sub-population~$\pop_{i-1}$. Ultimately, our method will produce neural networks which have trained with the hyperparameters of each sub-population in turn, starting with $\pop_n$ and ending with $\pop_1$. The fitness functions are designed to encourage this sequence of hyperparameter values to form a good schedule.

Evaluating fitness in parent sub-populations is the role of the evaluator workers. Throughout the experiment, an evaluator $\eta$ repeats the following procedure:

\begin{enumerate}
    \item Find the population member $\rho_{parent} \in \bigcup_{j=2}^P \pop_{j}$ and its sub-population $\pop_i$ which has least recently been evaluated and copy its weights: $\theta_\eta \leftarrow \theta_{\rho_{parent}}$. We say that $\eta$ is then \emph{assigned} to $\rho_{parent}$.
    \item Find the highest fitness member $\rho_{child}$ in $\pop_{i-1}$ and copy its hyperparameters: $h_\eta \leftarrow h_{\rho_{child}}$. We say $\rho_{child}$ is the \emph{target} of the evaluator.
    \item Start training from the newly assigned evaluator weights $\theta_\eta$ with the evaluator hyperparameters $h_\eta$ and regularly measure the objective function $\mathcal{Q}$. This sequence of evaluations of $\mathcal{Q}$ will generate a training curve which will be used to evaluate the fitness of $\rho_{parent}$. The larger the rate of improvement of the curve, the higher the fitness (detailed in Section \ref{sec:rate}).
    \item Continue training until either the evaluator's stopping or success criteria are reached (detailed in Section \ref{sec:eval-success}). If the evaluator's success criteria are reached, make the evaluator's target $\rho_{child}$ copy the evaluator's current weights: $\theta_{\rho_{child}} \leftarrow \theta_\eta$.
\end{enumerate}

Evaluators have two purposes: First, they evaluate the fitness of population members of parent sub-populations. Section \ref{sec:rate} describes how the training curve of an evaluator assigned to $\rho_{parent}$ is used to compute the fitness of $\rho_{parent}$. Second, evaluators make the link between sub-populations by propagating the trained weights of a parent sub-population into its child sub-population. Section \ref{sec:eval-success} describes the conditions for evaluator success where the evaluator's performance is considered greater than the one of its target $\rho_{child}$. When that occurs, $\rho_{child}$ copies the weights of the evaluator. This mechanism is how weights are propagated between sub-populations.

\subsection{Computing the fitness of parent sub-population members}
\label{sec:rate}
 
As in regular PBT, evolution in parent sub-populations is decided based on a fitness assigned to its members. This section discusses how the training curve of an evaluator that is assigned to a population member $\rho_{parent}$ is used to compute the fitness of $\rho_{parent}$. The intuition behind our method is that we can use the improvement rate of the training curve as a proxy for the quality of the original weights~$\theta{\rho_{parent}}$. Stated explicitly, the assumption that guides our method is as follows. 

\begin{assumption}
 Given two neural networks with similar performance and training with the same or similar hyperparameters, the network showing the fastest rate of improvement will lead to a better final performance.
 \label{ass:one}
\end{assumption}

Section \ref{sec:comparing} proposes a comparative method which exploits this assumption. Given two curves, it determines whether they are comparable and, if so, which of the two is improving faster. Section \ref{sec:fitness} will describe how this relative metric is turned into a global fitness.

\subsubsection{Comparing two curves}
\label{sec:comparing}

\begin{figure}
  \centering
  \includegraphics[width=0.8\linewidth]{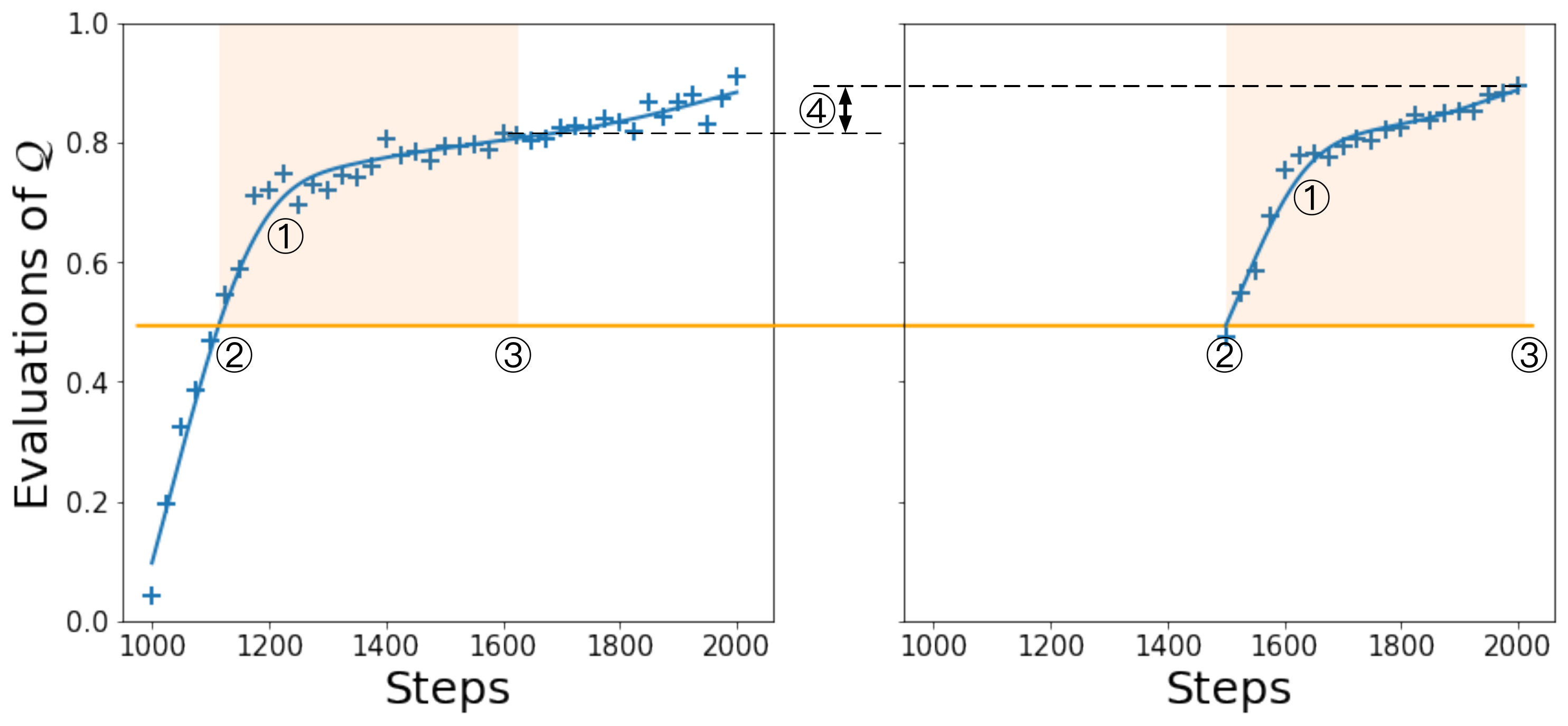}
  \caption{Example comparison of two curves. Although the left curve has a higher average improvement rate, our procedure find the right curve to be comparatively better. The circled numbers indicate the step in the \texttt{best\_score\_diff} procedure: 1) Smooth the curves, 2) Find the starts of the overlapping sections, 3) Find the length of the overlapping sections, and 4) Compare the best score within each overlapping section.}
  \label{fig:comparison}
\end{figure}

Each evaluator training curve consists of a series of points $[(x_1, y_1), (x_2, y_2), \dots , (x_n, y_n)]$ where the $x$s are regularly spaced training steps and $y$s are the evaluated value of the objective function $\mathcal{Q}$ at that step. $x_1$ is the step at which the evaluator copied the weights of its assigned member~$\rho_{parent}$ and $x_n$ is its current step. For example, Figure \ref{fig:comparison} shows a comparison between two curves as they both reach step 2000. The evaluator that produced the left curve copied the weights of its assigned member at step 1000, while the evaluator that produced the right curve did so at step 1500.

Given two evaluators $\eta$ and $\kappa$ and their respective training curves $(\mathbf{x}^{\eta}, \mathbf{y}^{\eta})$ and $(\mathbf{x}^{\kappa}, \mathbf{y}^{\kappa})$, we describe below a function \texttt{best\_score\_diff}($\eta$, $\kappa$) which is positive if $\eta$ is considered to be improving faster than $\kappa$ and negative if $\kappa$ is improving faster than $\eta$. It works by finding an \emph{overlapping section} where the two curves have similar performance and their rates of improvement can be compared. The function \texttt{best\_score\_diff} executes the following procedure:
\begin{enumerate}
    \item \textbf{Smooth the curves}. Fit a Gaussian Process (GP) \cite{gp} through each curve to smooth them. This is useful in contexts where $\mathcal{Q}$ fluctuates rapidly throughout training. We use a Mat\'{e}rn 5/2 kernel and use empirical Bayes to fit its hyperparameters. Let $\mathbf{\hat{y}}^\eta$ and $\mathbf{\hat{y}}^\kappa$ be the smoothed version of $\mathbf{y}^{\eta}$ and $\mathbf{y}^{\kappa}$ respectively.
    \item \textbf{Find the starts of the overlapping sections}. Find the curve whose GP fit has the highest starting performance by comparing $\hat{y}^{\eta}_1$ and $\hat{y}^{\kappa}_1$. Then, find the first point when the other curve's GP fit reaches that performance for the first time. That is, say $\hat{y}^\eta_1<\hat{y}^\kappa_1$, we find the smallest $r$ such that $\hat{y}^\eta_r \geq \hat{y}^\kappa_1$. The overlapping section of $(\mathbf{x}^\eta, \mathbf{y}^\eta)$ will start at $(x^\eta_r, y^\eta_r)$, and the one of $(\mathbf{x}^\kappa, \mathbf{y}^\kappa)$ at $(x^\kappa_1, y^\kappa_1)$. For example, in Figure \ref{fig:comparison}, the right curve starts higher and so its overlapping section starts from its beginning. The left curve first reaches a similar performance at around 1150 steps, which is where its overlapping section starts.
    \item \textbf{Find the length of the overlapping sections}. Find the maximum number of steps for which both curves have evaluations of $\mathcal{Q}$ data past their starting point. That is, say the overlapping sections start at $(x^\eta_r, y^\eta_r)$ and $(x^\kappa_s, y^\kappa_s)$, then the length of the overlapping sections is $n = \max(\texttt{len}(\mathbf{x}^\eta)-r, \texttt{len}(\mathbf{x}^\kappa)-s)$. In Figure \ref{fig:comparison}, the right curve has fewer steps past its starting point and therefore constrains the length of the overlapping section. 
    \item \textbf{Compare the best score within each overlapping section}. Find the highest recorded evaluation of $\mathcal{Q}$ within each curve's overlapping section and compute their difference: $\texttt{best\_score\_diff}(\eta, \kappa) = \max{\mathbf{y}^\eta_{r:r+n}} - \max{\mathbf{y}^\kappa_{s:s+n}}$. 
\end{enumerate}

Note that the smoothed curves obtained in Step 1 are only used in the second step. In general, we prefer relying on the data of the original training curve for simplicity and robustness. However, we find smoothing the curves is necessary to obtain reasonable overlapping sections.

A special case occurs when two evaluators do not have an overlapping section because one curve is strictly higher than the other. This is an important case to tackle: it frequently occurs when poor hyperparameter values are explored. These will often lead to degenerate neural network weights and flat training curves with a minimal score. In order to respond to these contexts appropriately we make a second assumption.

\begin{assumption}
 When training with a fixed set of hyperparameters, the training curves of neural networks are concave. 
\end{assumption}

We use this assumption in the following way. Say the training curve of $\eta$ is strictly higher than the one of  $\kappa$: $\min(\mathbf{y}^\eta) > \max(\mathbf{y}^\kappa)$. Our method tries to penalise the training curve of $\eta$ by a constant amount to see if, despite being higher, it is also improving faster. That is, we say $\texttt{penalised}_\delta(\eta)$ has the training curve $(\mathbf{x}, \mathbf{y} - \delta)$ and check if $\texttt{best\_score\_diff}(\texttt{penalised}_\delta(\eta), \kappa) > 0$. We try multiple values for $\delta$ in the range $[\min(\mathbf{y}^\eta) - \max(\mathbf{y}^\kappa), \min(\mathbf{y}^\eta) - \min(\mathbf{y}^\kappa)]$. If some values of $\delta$ lead to a positive score, we make \texttt{best\_score\_diff}($\eta$, $\kappa$) return the highest found score. Otherwise, we set \texttt{best\_score\_diff}($\eta$, $\kappa$) = 0.

\subsubsection{Computing the fitness of members of parent sub-population}
\label{sec:fitness}

We now show how to compute the fitness of the members of a parent sub-population $\pop_{i}$. First, we find the set $\Phi$ of members in $\pop_{i}$ which currently have an assigned evaluator. Note that members never have more than a single evaluator assigned to them. Members that have not yet been assigned an evaluator do not have a fitness and are exempt from evolution.

Second, for each member in this set, we compare its evaluator curve to the evaluator curve of each other member in the set. We then sum the computed \texttt{best\_score\_diff} over all comparisons. Call $\eta_{\rho}$ the evaluator of $\rho$, we set:
$$\texttt{fitness}(\rho_j) = \sum_{\rho_k \in \Phi} \texttt{best\_score\_diff}(\eta_{\rho_j}, \eta_{\rho_k}).$$

\subsection{Evaluator stopping and success criteria}
\label{sec:eval-success}

Evaluators have two purposes. The first one is producing the training curves to evaluate the members of parent sub-populations as described in the previous section. The second is providing the link between sub-populations which we present here.

Our method keeps evaluators training as long as it is plausible that they will reach a higher performance than the one of their target $\rho_{child}$. If it looks like this will not happen, we stop the evaluator, making it available to evaluate other population members. In order to make robust decisions, our method uses a notion of statistical significance which we present below.

\textbf{Testing for statistical significance.} To see if an evaluator $\eta$ is likely to outperform its target $\rho_{child}$, we compare their training curves. Our goal is to establish whether the training curve of $\eta$ is improving statistically significantly faster than $\rho_{child}$. According to Assumption \ref{ass:one}, this would imply the neural network weights held by $\eta$ will result in better long-term performance. As before, the training curve of $\eta$ starts from when it copied the weights of its assigned member $\rho_{parent}$. The training curve of $\rho_{child}$ starts from when it last copied the weights of another member via an exploit-and-explore step.

Given the training curves $(\mathbf{x}^\eta, \mathbf{y}^\eta)$ of $\eta$ and $(\mathbf{x}^{\rho_{child}}, \mathbf{y}^{\rho_{child}})$ of $\rho_{child}$, we start by extracting overlapping regions $(\mathbf{x}^\eta_{r:r+n}, \mathbf{y}^\eta_{r:r+n})$ and $(\mathbf{x}^{\rho_{child}}_{s:s+n}, \mathbf{y}^{\rho_{child}}_{s:s+n})$ as described in Section \ref{sec:comparing}. We then do a pairwise comparison of each of the points in $\mathbf{y}^\eta_{r:r+n}$ and $\mathbf{y}^{\rho_{child}}_{s:s+n}$: $y^\eta_{r+1} > y^{\rho_{child}}_{s+1}, \dots, y^\eta_{r+n} > y^{\rho_{child}}_{s+n}$. Let $k$ be the number of times $\mathbf{y}^\eta$ was higher than $\mathbf{y}^{\rho_{child}}$. We run a binomial test with $k$ successes and $n$ trials and compute the $p$-value. We call this quantity $\texttt{binom\_test}(\eta, \rho_{child})$.

\textbf{Stopping criteria.} At regular intervals, an evaluator will compare its training curve with the one of its target $\rho_{child}$ using the procedure described above. An evaluator that has been training for $T$ steps will stop if either of the following criteria is met:
\begin{itemize}
    \item The evaluator's curve does not overlap with $\rho_{child}$'s curve and $T$ is greater than a hyperparameter \texttt{max\_eval\_steps}, or
    \item The evaluator's curve does overlap with $\rho_{child}$'s curve but $\texttt{binom\_test}(\eta, \rho_{child})$ is greater than $p_{stat} + max(0,  1 -T/\texttt{max\_eval\_steps})$ where $p_{stat}$ is a hyperparameter of our method. We use $p_{stat} = 0.01$ throughout our experiments.
\end{itemize}

\textbf{Success criteria.} If $\eta$ succeeds, meaning it is deemed better than $\rho_{child}$, $\rho_{child}$ will discard its current neural network weights and copy the one of the evaluator: $\theta_{\rho_{child}} \leftarrow \theta_\eta$. The conditions for evaluator success are: 1) $\texttt{best\_score\_diff}(\eta, \rho_{child}) > 0$, and 2) $\texttt{binom\_test}(\eta, \rho_{child}) < p_{stat}$.

Finally, an evaluator will stop training and look for a new task if either the member it is assigned to $\rho_{parent}$ or its target $\rho_{child}$ loses an evolution event and executes an exploit-and-explore step. When choosing a new member to evaluate, the evaluator will pick the one which has been training for the longest since either it was evaluated or lost an evolution event. 

In some cases, we find it beneficial to use a hyperparameter $\texttt{min\_steps\_before\_eval}_i$ per parent sub-population $\pop_i$ which determines the minimum number of steps before its members can be evaluated. This allows us to make sure that its child population $\pop_{i-1}$ has converged on a set of hyperparameters. This way, the evaluators generating training curves for the members of $\pop_i$ will all be using similar hyperparameters. We tune $\texttt{min\_steps\_before\_eval}_i$ by inspecting the number of steps needed for $\pop_{i-1}$ to converge on a narrow set of hyperparameter values.

\section{Experiments}
\label{sec:experiments}

Our experiments focus on hyperparameter optimisation methods that are able to complete within the wall clock time of a single learning process. We compare FIRE PBT to PBT and random hyperparameter search (RS) on an image classification task and a reinforcement learning (RL) task.

\textbf{Selecting FIRE PBT hyperparameters.} The main hyperparameters to choose in FIRE PBT are the number of sub-populations and the distribution of workers into the different sub-populations and evaluator set. Within an experiment, we always make all sub-populations be the same size. That size is a function of the difficulty of the optimisation problem. We find 8 to be a good size for our image classification experiments where learning is very stable and use 18 in our reinforcement learning experiments where training is more stochastic. We set the number of evaluators to be three quarters of the total size of the parent sub-populations. This is because, at any one time, roughly one quarter of members will have just performed an explore-and-exploit step and so should train for some time before being evaluated. 

\subsection{Image classification}

We compare FIRE PBT with PBT and RS on the classification task of training a ResNet-50 \cite{resnet} on the ImageNet dataset \cite{imagenet} with a batch size of 1024. We optimise the learning rate hyperparameter.

The ImageNet dataset consists of \textasciitilde1.28 million training images and 50,000 validation images. To avoid overfitting the validation set, we extract 10,000 samples at random from the training set and use them for measuring the objective function $\mathcal{Q}$. For clarity, we refer to these extracted samples as the validation set, and the original 50,000 validation images as the test set.

For each method, we run five experiments each with 50 workers training on the truncated training set. To show the impact of population size, we also run FIRE PBT experiments with 22 workers and 36 workers. For each of these experiments, we find the highest recorded validation score throughout training. We report this score as well as the test score that was achieved by the same neural network at that same point in training. Finally, we retrace the hyperparameter schedule that lead to this score, including the number of steps trained with each hyperparameter. We then train a new neural network, from scratch, following this hyperparameter schedule and training on the entire training set. We record the final test performance. This final step is so that our scores can be compared with previously published results produced in this regime.

\subsubsection{Experimental details}

In our random search experiments, we follow the already tuned hyperparameter schedule proposed by Goyal et al. \cite{goyal}. Their procedure uses a reference learning rate $\lambda \frac{batch\_size}{256}$ with $\lambda =$ \num{1e-1}. During the first 5 epochs of training, the schedule linearly increases the learning rate from 0 to the reference learning rate. It then keeps it constant, reducing it by 1/10 at the 30-th, 60-th and 80-th epoch. In our experiments, we keep this relative schedule and set each of the 50 workers to randomly sample $\lambda$ log-uniformly between \num{1e-2} and \num{3e-1} at the start of the experiment. 

We use the same hyperparameters for both regular PBT and FIRE PBT's within sub-population PBT:

\begin{table*}
\centering
 \begin{tabular}{ l   c   c  c } 
 \hline
 & Top validation acc. & Matching test acc. & Schedule replay test acc.  \\ 
 \hline
  RS Hand-tuned schedule& $\mathbf{80.49} \pm 0.13$ & $76.04 \pm 0.16$ & $\mathbf{76.60} \pm 0.16$ \\ 
 PBT &  $75.78 \pm 0.78$  &  $70.91 \pm 0.78$ & $71.69 \pm 0.71$ \\ 
 FIRE PBT (22) &  $80.07 \pm 0.31$  & $76.17 \pm 0.08$  & $76.04 \pm 0.08$ \\ 
 FIRE PBT (36) &  $80.36 \pm 0.35$ & $\mathbf{76.43} \pm 0.24$ & $76.36 \pm 0.29$ \\ 
 FIRE PBT (50) & $\mathbf{80.59} \pm 0.31$   & $76.06 \pm 0.51$ &  $\mathbf{76.51} \pm 0.34$ \\ 
 \hline
 \end{tabular}
\caption{ImageNet scores. Shows mean and standard deviation of the top validation accuracy scored throughout the experiment, the test accuracy scored at the same point in training, and the test accuracy obtained by replaying the hyperparameter schedule found and training on the full training set.}
\label{table:imagenet_results}
\end{table*}

\textbf{Hyperparameters} We optimise the hyperparameter $\lambda$ which determines the learning rate of $\lambda \frac{batch\_size}{256}$. As with our random search experiments, we sample the initial value of $\lambda$ log-uniformly between \num{1e-2} and \num{3e-1}.

\textbf{Objective function} We evaluate a model by computing its top-1 accuracy on the validation set. In order to generate detailed training curves for FIRE PBT, we evaluate this every 200 training steps.

\textbf{Ready} A member of the population is deemed ready to exploit and explore every 2400 steps. Similarly, in FIRE PBT, evaluators check their stopping and success criteria every 2400 steps.

\textbf{Exploit} We use a truncation selector: If a population member has a fitness in the bottom 25\% of the population, it copies the neural network weights and hyperparameters of a random member in the top 25\% of the population.

\textbf{Explore} We multiply $\lambda$ by a value sampled at uniform random from [0.5, 0.8, 1.25, 2.0].

In our PBT experiments, the 50 workers form a single population. In FIRE PBT, our experiments with 22, 36 and 50 workers use two, three and four sub-populations respectively. We set each sub-population to be of size 8 and use the remaining workers as evaluators. We set \texttt{max\_eval\_steps} to 7200 steps. We find that using the $\texttt{min\_steps\_before\_eval}$ hyperparameter is unnecessary as the sub-populations quickly converge to reasonable hyperparameter values and set it to 0 for all sub-populations. Both PBT and FIRE PBT use a synchronous implementation which will wait for all workers to have reached the next evolution step before making any decision. 

\subsubsection{Results}
\label{sec:imagenet_res}

\begin{figure}[t]
  \centering
  
     \begin{subfigure}[b]{0.49\textwidth}
         \centering
         \includegraphics[width=\textwidth]{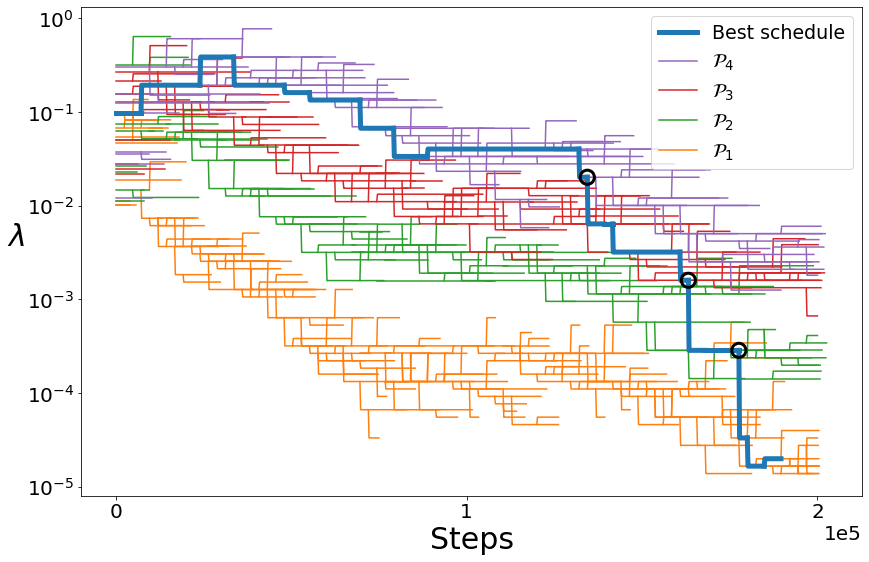}
         \caption{Distribution of the $\lambda$ parameter (learning rate up to a linear scale) of the four sub-populations throughout a FIRE PBT experiment.}
         \label{fig:imagenet_hyperstt}
     \end{subfigure}
     \hfill
     \begin{subfigure}[b]{0.49\textwidth}
         \centering
         \includegraphics[width=\textwidth]{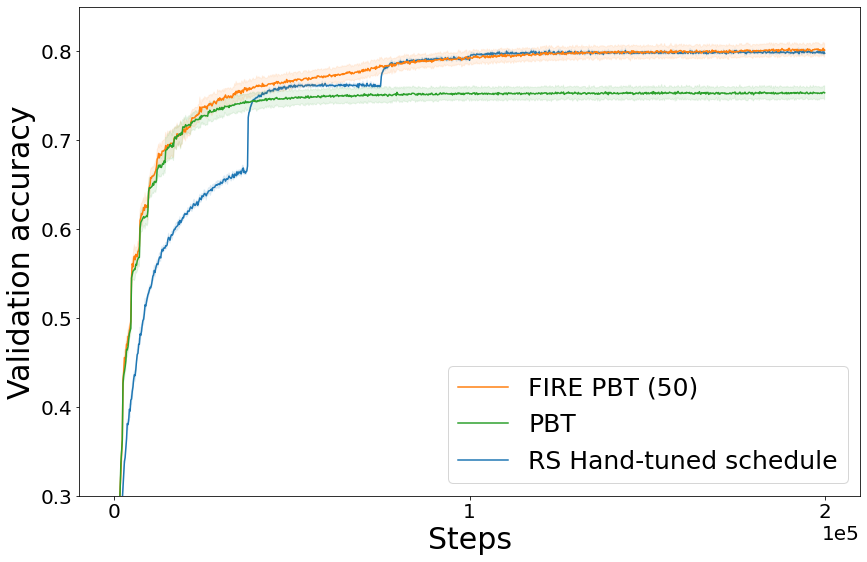}
         \caption{Distribution across experiments of the population's best validation accuracy scored by a worker at each step.}
         \label{fig:besttt}
     \end{subfigure}
  \caption{Evolution of the hyperparameters (left) and validation scores (right) throughout ImageNet training.}
\end{figure}

Table \ref{table:imagenet_results} shows for each method: the mean and standard deviation of the highest validation score, the test score that was achieved at the same point, and the test score that was achieved by replaying the same hyperparameter schedule and training on the full training set. FIRE PBT significantly outperforms PBT and achieves results similar to those of Random Search with a hand-tuned schedule.

Figure \ref{fig:imagenet_hyperstt} shows the hyperparameter distribution of the four sub-populations for one of our FIRE PBT experiments as well as the hyperparameter schedule that lead to the best validation score. The black circles indicate the points at which evaluators copied the neural network weights of their assigned member, leading to evaluator success. All other hyperparameter mutations that are shown are the result of an exploit-and-explore step.

One of the advantages of FIRE PBT is that at any time throughout the experiment, the population members of $\pop_1$ provide a current best effort performance. Figure \ref{fig:besttt} shows the best validation score achieved at any point in our experiments. We can see that FIRE PBT rapidly reaches high accuracy without compromising the long-term performance.

\textbf{Compute costs.} The cost of a single end-to-end ImageNet training in our setup including the frequent evaluations is approximately 4 TPUv3-core-days. This brings the cost of an experiment with 50 workers to 200 TPUv3-core-days.

\subsection{Reinforcement learning}
\label{sec:rl_exps}

We compare FIRE PBT with RS and PBT on the task of optimising the hyperparameters of a reinforcement learning agent training with a V-MPO loss \cite{vmpo}. We replicate the setting of the original V-MPO experiments on the OpenAI Gym tasks of Ant-v1 and Humanoid-v1 \cite{openaigym}, including all hyperparameters. These experiments sweep over two hyperparameters showing robustness to their values: $\epsilon_{\alpha_\mu}$ and $\epsilon_{\alpha_\Sigma}$, which bound the rate of change of the policy, are sampled log-uniformly in the ranges [\num{5e-3}, \num{1e-2}] and [\num{5e-6}, \num{5e-5}] respectively. As their values have little impact on performance, we fix them to \num{7.5e-3} and \num{1e-5} respectively.

For each method, we run three experiments each with 50 workers. All three methods optimise the same hyperparameters and use the same initial sampling distributions. We also use the same hyperparameters for both PBT and FIRE PBT's within sub-population PBT. In FIRE PBT, we split the workers into two sub-populations of size 18 and use the remaining 14 workers as evaluators. Sub-population $\pop_2$ uses \texttt{min\_steps\_before\_eval} of \num{1e9} actor steps and \texttt{max\_eval\_steps} of \num{6e8} actor steps. We let agents train for \num{6e9} actor steps.

\textbf{Hyperparameters} We optimise the learning rate and the $\epsilon_\eta$ hyperparameter which sets the size of a trust region. We initially sample the learning rate log-uniformly between \num{1e-5} and \num{1e-3} and $\epsilon_\eta$ log-uniformly between \num{5e-3} and \num{2e-2}. 

\textbf{Objective function} We evaluate the current model by computing the average episode return.

\textbf{Ready} A member of the population is deemed ready to exploit and explore every \num{6e7} actor steps. Similarly, in FIRE PBT, evaluators check their stopping and success criteria every \num{6e7} steps.

\textbf{Exploit} We use a truncation selector: If a population member has a fitness in the bottom 25\% of the population it copies the neural network weights and hyperparameters of a random member in the top 25\% of the population.

\textbf{Explore} We perturb each of the hyperparameters independently. The learning rate is multiplied by a random value selected from [0.5, 0.8, 1.25, 2.0]. Likewise,  $\epsilon_\eta$ is multiplied by a random value selected from [0.8, 0.95, 1.05, 1.2].

\begin{figure}
  \centering
  \includegraphics[width=\linewidth]{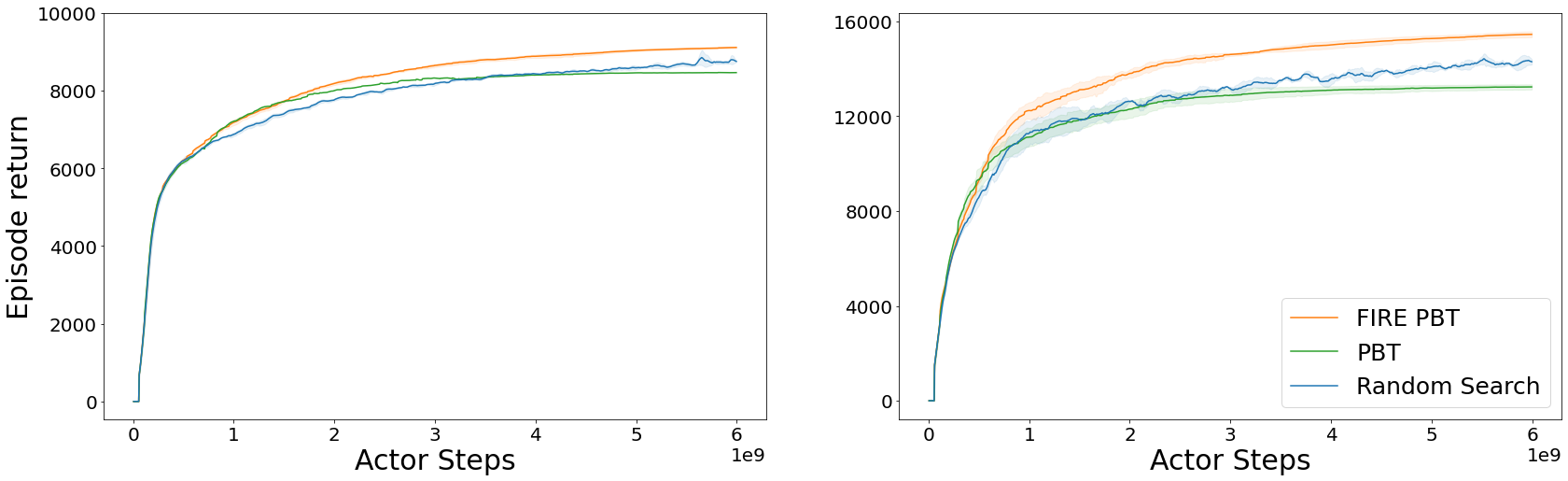}
  \caption{Distribution across experiments of the top average episode return scored by a worker at each step on Ant-v1 (left) and Humanoid-v1 (right).}
  \label{fig:vmpo}
\end{figure}

Figure \ref{fig:vmpo} shows the distribution across experiments of the top average episode return achieved by one of the population's workers throughout training. FIRE PBT shows faster learning and higher performance than PBT and RS.

\textbf{Compute costs.} The cost of training a single OpenAI Gym agent in our setup is approximately 3.2 TPUv2-core-days and 3.2k CPUs-days. This brings the cost of an experiment with 50 workers to 160 TPUv2-core-days and 160k CPUs-days.

\section{Related work}
\label{sec:relatedwork}
The field of AutoML \cite{automl} has grown rapidly in recent years due to the demand for machine learning methods that work out of the box. Most of these methods focus on sequential optimisation, where the result of a training run is used to inform the next iteration of hyperparameter search. In this domain, Bayesian optimisation has seen a large amount of attention \cite{snoek2012practical, hutter2011sequential, bergstra2011algorithms}. Some approaches use parallelism and evaluate multiple hyperparameters at once to reduce the number of iterations~\cite{shah2015parallel, gonzalez2016batch, wu2016parallel}. Compared with these approaches, PBT and FIRE PBT have the advantage of completing within the wall clock time of a single learning process. 

Some work in this area models the training curves of models to predict how they will do in the future \cite{swersky2014freeze, 10.5555/2832581.2832731, DBLP:conf/iclr/KleinFSH17}. In concept, this is similar to the faster improvement rate approach presented in this paper. However, these works typically use this approach to compare the training curves of networks training with different hyperparameters, whereas we restrict our method to comparing networks that are training with similar hyperparameters. 

Hyperband \cite{li2016hyperband} is a purely parallel approach which models hyperparameter selection as an infinite-armed bandit. However, the initial amount of parallelism required by the method is large and hence a practical implementation will start by sequentially evaluating independent models.

An important fraction of the work in AutoML focuses on the hyperparameters that guide the architecture of the neural network \cite{archsearch}. It is an open research question how such hyperparameters should be included in the PBT or FIRE PBT framework as it would involve altering a network's architecture during training. 

\section{Conclusion}
\label{sec:conclusion}

We have presented Faster Improvement Rate PBT, an extension of regular PBT, which tackles the cases in which greedy hyperparameter mutations lead to poor long-term performance. We have shown the effectiveness of FIRE PBT on the distinct domains of image classification and reinforcement learning. Our experiments showed that FIRE PBT was able to find sensible hyperparameter schedules which matched the performance of hand-tuned ones and outperformed static schedules. We hope that the use of FIRE PBT will become a powerful platform for researchers and practitioners to explore new methods.

\bibliography{main}

\end{document}